# AI Agents in Drug Discovery

Srijit Seal[1,2,†,*], Dinh Long Huynh[3,†], Moudather Chelbi[4], Sara Khosravi[1,5], Ankur Kumar[1], Mattson Thieme[6], Isaac Wilks[6], Mark Davies[7], Jessica Mustali[8], Yannick Sun[8], Nick Edwards[9], Daniil Boiko[10], Andrei Tyrin[10], Douglas W. Selinger[11], Ayaan Parikh[12], Rahul Vijayan[12], Shoman Kasbekar[13,14], Dylan Reid[15], Andreas Bender[16,17,2], Ola Spjuth[3,18,*]

**Affiliations:**
[1] Broad Institute of MIT and Harvard, Cambridge, MA 02142, US
[2] University of Cambridge, Cambridge CB2 1EW, UK
[3] Department of Pharmaceutical Biosciences and Science for Life Laboratory, Uppsala University, Uppsala, 75237, Sweden
[4] Augmented Nature, Berlin, 10319, Germany
[5] University of California, Los Angeles (UCLA), Los Angeles, CA, US
[6] Human Chemical, San Francisco, CA 94121, US
[7] Kiin Bio, London, EC2A 1NT, United Kingdom
[8] Misogi Labs, San Francisco, CA 94104, US
[9] Happy Potato, Inc., Seattle, WA 98104, US
[10] Onepot AI, Inc., South San Francisco, CA 94080, US
[11] Plex Research, Cambridge, MA, US
[12] Convexia, San Francisco, CA 94104, US
[13] Kepler AI, San Francisco, CA, US
[14] Nome Bio, San Diego, CA, US
[15] Zetta Venture Partners, San Francisco, CA,US
[16] College of Medicine and Health Sciences, Khalifa University of Science and Technology, Abu Dhabi, UAE
[17] STAR-UBB Institute, Babeş-Bolyai University, Cluj-Napoca, Romania
[18] Pixl Bio AB Uppsala, SE-75239, Sweden

*†Equal contribution;*
*\*Corresponding authors: seal@broad.harvard.edu (S.S.), ola.spjuth@uu.se (O.S.)*

**Abstract**

Artificial intelligence (AI) agents are emerging as transformative tools in drug discovery, with the ability to autonomously reason, act, and learn through complicated research workflows. Building on large language models (LLMs) coupled with perception, computation, action, and memory tools, these agentic AI systems could integrate diverse biomedical data, execute tasks, carry out experiments via robotic platforms, and iteratively refine hypotheses in closed loops. We provide a conceptual and technical overview of agentic AI architectures, ranging from ReAct and Reflection to Supervisor and Swarm systems, and illustrate their applications across key stages of drug discovery, including literature synthesis, toxicity prediction, automated protocol generation, small-molecule synthesis, drug repurposing, and end-to-end decision-making. To our knowledge, this represents the first comprehensive work to present real-world implementations and quantifiable impacts of agentic AI systems deployed in operational drug discovery settings. Early implementations demonstrate substantial gains in speed, reproducibility, and scalability, compressing workflows that once took months into hours while maintaining scientific traceability. We discuss the current challenges related to data heterogeneity, system reliability, privacy, and benchmarking, and outline future directions towards technology in support of science and translation.

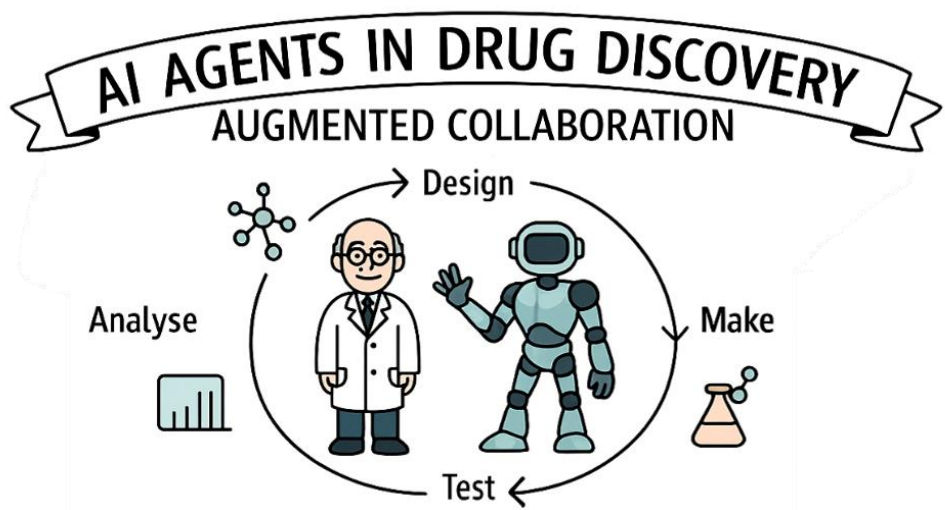

# 1. Introduction

Drug discovery is a long, expensive, and high-risk process that spans from identifying new biological targets to delivering safe and effective medicines to patients[1]. At a general and simplified level, it follows a sequence of connected stages: target identification and validation, hit discovery, lead optimization through iterative Design-Make-Test-Analyze (DMTA) cycles, preclinical safety assessment, and finally, clinical developmen[2–5]. Each stage produces large amounts of data, ranging from omics and imaging to assay readouts, pharmacokinetic studies, and clinical outcomes[6]. Despite technological advances, such as non-animal new approach methodologies[7] and the use of machine learning models, complex decision-making at key junctures is still largely manual, repetitive, and rely on cross-disciplinary teams to search, collate, and interpret scatter evidence[8]. This reliance on fragmented workflows contributes significantly to the rising cost, longer timelines, and high attrition rates that characterize drug R&D. A major bottleneck persists in communication of results, repeated planning meetings, presenting results, discussions, and deliberations to make key decisions.

Computational methods, and current AI in an advanced form, is already influencing methods used in drug discovery in multiple ways. Predictive AI in the form of machine learning has enabled quantitative structure-activity relationship (QSAR) models that predict compound properties to be lifted to a quantitatively, and probably also qualitatively new level[9–11]. Generative AI has made it possible to design new chemical structures with desired properties, and large language models (LLMs) are starting to structure machine-readable data in order to propose hypotheses[12–16] (**Figure 1**). Still, these paradigms share a common limitation: they comprise passive implementations requiring human actions for repetitive tasks such as to prepare inputs, and interpret outputs[16,17]. They cannot autonomously orchestrate the multi-step reasoning and decision-making required across the complex, interdependent processes of drug discovery.

A new wave of systems, collectively termed agentic AI, offers the potential to overcome these limitations (**Figure 1**). Agentic AI builds on the recent advancements in LLMs' reasoning capabilities but couples them with external tools, memory, and data sources, enabling systems that can 'think', 'act', 'observe', and 'reflect' in iterative loops. Rather than serving as a single predictive or generative model, these agents function as adaptive independent components that can act together in retrieving literature, triaging compounds, predicting toxicity endpoints such as liver or cardiotoxicity[18], planning experiments, and even interfacing with lab automation[19].

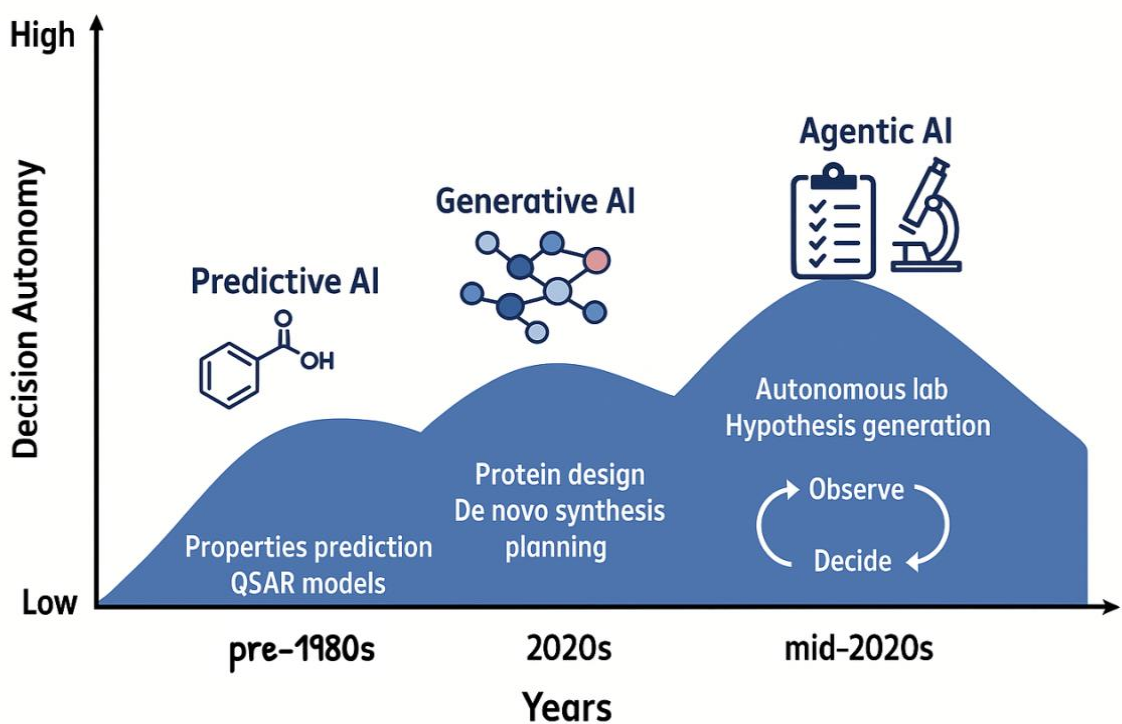

**Figure 1.** The Three Evolutionary Waves of Artificial Intelligence: From Predictive Models to Generative and Agentic Systems.

This work aims to place agentic AI directly in the context of the drug discovery pipeline. We first outline the conceptual foundations and system architectures of agentic AI. We then examine applications across key stages of drug R&D, from target discovery and DMTA cycles to preclinical safety and drug repurposing, highlighting case studies where agents are already beginning to reshape practice. Finally, we discuss important challenges, including data heterogeneity, infrastructure integration, and system reliability, and propose future directions.

## 2. Agentic AI: Concepts and System Design Principles

### 2.1. Agentic AI - the new AI application wave for drug discovery

The advance of predictive and generative AI in drug discovery, especially Large Language Models (LLMs), has paved the way for the evolution of agentic AI. This advancement represents an extension of engineering efforts to coordinate prior predictive and generative AI models to autonomously perform various tasks in drug discovery end-to-end. At the heart of these agentic systems are LLMs with 'reasoning' capacity[5,19–22], the extent of which is still subject to debate[23,24]. A stand-alone LLM is a generative model, which cannot interact with the real world to perform meaningful actions[25] or access external data beyond its training set, i.e., domain-specific data or proprietary data. To acquire such capabilities, the LLM must integrate with external tools[20,21,26]. Here, tools refer to any external components integrated with an LLM that can be called to perform specific tasks. There are four types of common agent tools in drug discovery based on their functionality (**Figure 2**). Perception tools serve as the system's augment layer, gathering information from both structured and unstructured biomedical databases, such as ChEMBL, PubChem, STRING, and Reactome[27–30]. Computation tools enable predictions, simulations, data analysis, or other computational actions in the drug discovery field. These tools often serve as wrappers for pre-trained models, such as AlphaFold[31], or data processing pipelines, like NextFlow[32]. Execution of these tools usually occurs on local computers or on cloud high-performance computing platforms in order to enable the agentic systems to handle large-scale data. Action tools provide the agentic system with the ability to act in the real world. For example, the agentic system connects to robotic pipetting, automated cell-based assays, or next-generation sequencing library preparation to perform physical execution. These action tools robustly close the loop between *in silico* design and empirical validation. Memory tools maintain persistence across tasks and sessions by storing, retrieving, compressing, and updating the agent's working knowledge. In drug discovery contexts, memory databases might capture SAR patterns, accumulated toxicity findings, and other high-value knowledge that are repeatedly reused. This memory layer provides context persistence across multi-step reasoning, which enables the system to refine hypotheses over time. Together, these 4 types of tools enable agentic AI to implement end-to-end pipelines in drug discovery: Perception takes evidence and forms a hypothesis, Computation transforms that hypothesis into predictions, Action tests those predictions experimentally, and Memory retains what is learned from the previous cycles.

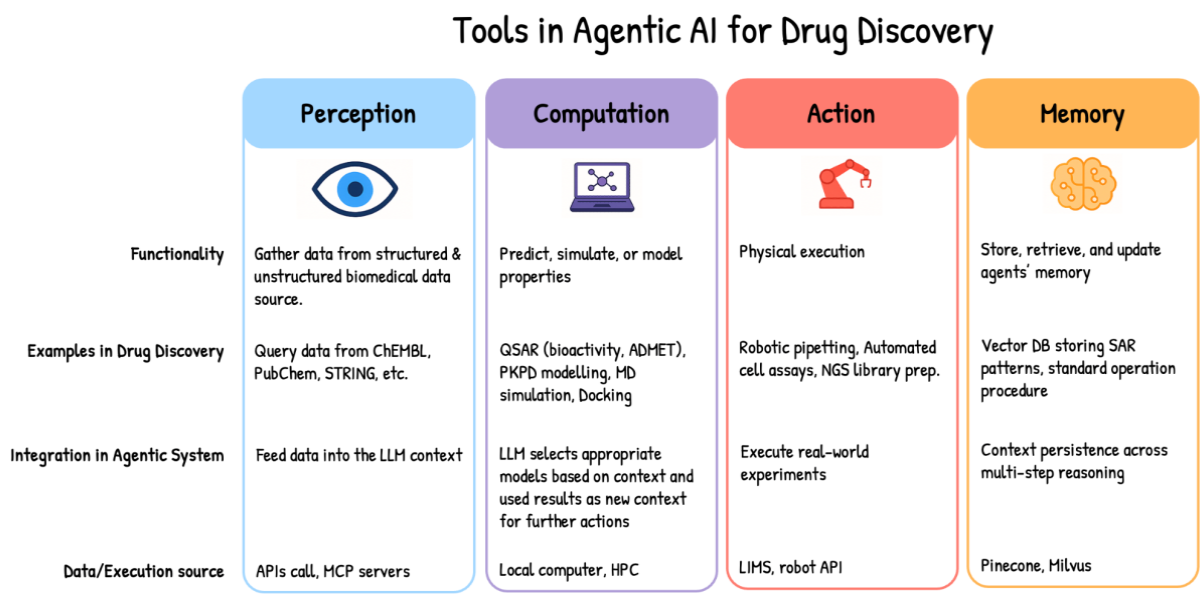

**Figure 2. Four typical tool types in agentic AI for drug discovery.** Perception tools enable data collection and integration, Computation tools perform analytical and predictive modelling, Action

tools execute real-world experimental operations, and Memory tools manage knowledge storage, retrieval, and continuous learning.

## 2.2. Agentic AI architectures

To better understand the capabilities of agentic AI, it is helpful to examine several established system architectures. The most basic agentic architecture is the ReAct agent, which stands for Reasoning-Acting[33] (**Figure 3a**). This design enables the LLM to dynamically select and execute the necessary tools when receiving tasks. This architecture also enables an iterative loop between the reasoning and acting processes. The LLM actively determines when the process will terminate with minimum human intervention. This enhances the flexibility and adaptability of the agentic system, making it more akin to the style of research that involves thinking critically. Concomitantly, the iterative loop process mimics the DMTA cycle in drug discovery, strengthening the hypothesis with each iteration.

Another architecture is the reflection agent (**Figure 3b**), in which multiple LLMs are connected, enabling them to communicate with each other and critique one another's reasoning[34]. This system mimics working in pairs or group situations, where medicinal chemists sit together to discuss the synthesis plan for compounds. Therefore, this architecture is especially effective for tasks that require discussion and strategic planning. For example, it can be applied to plan multi-step synthesis routes or design experimental workflows for high-throughput screening.

Along with the advancement in the field, multi-agent architectures are also gaining traction, with the emergence of supervisor architectures (**Figure 3c**). This architecture is modelled on the way a research team is organized. We have one supervisor agent who focuses on task delegation, and many expert agents who deeply reason and act on a relevant specific task[19,35]. When receiving a task, the supervisor decomposes it into sub-tasks that are suitable for each specialized agent. Then, the supervisor delegates tasks to the sub-agents to execute and collect results. This process also enables an iterative loop similar to the ReAct agent, which makes the supervisor architecture goal-oriented.

However, in a supervisor architecture, every communication must go through the supervisor because specialized sub-agents only connect to the supervisor, not to each other. This repeated communication with the supervisor lengthens the process's context, which is a limitation for LLM with restricted context-window length. As a solution for this issue, the swarm architecture (**Figure 3d**) aims to decentralize the role of the supervisor agent. In swarm architecture, each agent has a connection to every other agent, facilitating the collaboration between them[36].

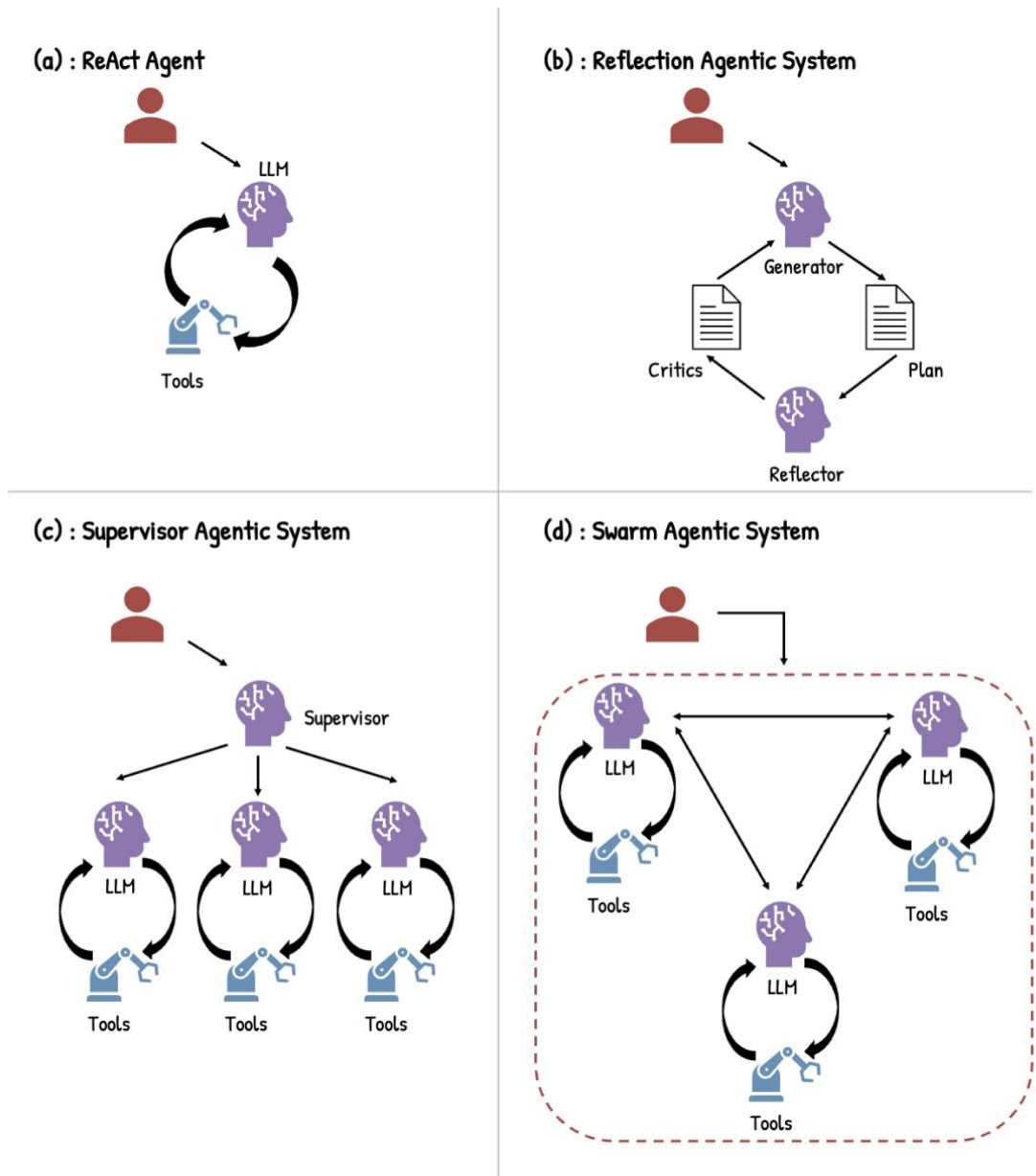

**Figure 3. Architectures of Agentic AI Systems.** (a) ReAct Agent: A reasoning-acting loop where a large language model (LLM) dynamically selects and invokes tools. (b) Reflection Agentic System: A multi-LLM setup in which a generator produces plans, a reflector critiques them, and iterations improve strategy. (c) Supervisor Agentic System: A hierarchical multi-agent architecture with a supervisor LLM delegating subtasks to specialized LLM agents, each capable of reasoning and tool use. (d) Swarm Agentic System: A decentralized multi-agent system with all agents connecting to each other, each capable of reasoning and tool use.

The similarity between supervisor and swarm architectures lies in their nature as multi-agent systems, where numerous agents communicate with each other. Agents can exchange data and insights through communication protocols. IBM's Agent Communication Protocol (ACP) and Google's Agent2Agent (A2A) which recently merged under the Linux Foundation, aim to create a standardized interface for agents to exchange data independently of the underlying framework or location which enables agents to collaborate across departments of the same organization and across companies and labs all over the world[37,38].

Overall, these designs represent just the beginning of a rapidly evolving field. As we continue to explore new effective architecture of agentic systems, many novel and domain-specific suggestions will emerge. The design space is wide open: from closely coupled experiment-agent systems to entirely autonomous research platforms. **Table 1** highlights the key features, advantages/disadvantages, and recommended tasks for each type of agentic system in the context of drug discovery.

**Table 1.** Agentic AI architectures and drug discovery task suggestions for each architecture.

| Agent Architecture | Key Features | Strengths | Limitations | Potential Drug Discovery Tasks |
|---|---|---|---|---|
| **ReAct** | Interleaved reasoning and acting, dynamic tool calls | High adaptability, minimal setup | May hallucinate tool use | Literature triage, SAR exploration |
| **Reflection** | Multiple LLMs critiquing each other | Improves reasoning quality | Slower, more computationally intensive | Multi-step synthesis route planning |
| **Supervisor** | Hierarchical orchestration with specialist agents | Handles complex, multi-domain tasks | Needs robust task decomposition | Autonomous HTS campaign management |
| **Swarm** | Decentralized peer agents | Scales across institutions | Coordination overhead | Federated multi-site toxicology data integration |

**2.3. Memory - Effective components for agents to learn**

An agent's memory gives it the ability to learn from experience, maintain coherent interactions, and accomplish complex tasks. It bridges the gap between internal knowledge and dynamic external data. There are two types of agents' memories (1) short-term which reflects the working space for immediate context and (2) long-term for persistent knowledge. In drug discovery, an agent's memory allows it to integrate continuously generated data. For instance, the agent can connect a finding from a new experimental result with internal memory of scientific publications to propose a novel hypothesis. It can instantly refine structural features of a new compound based on hurdles faced in past discovery programs. This active synthesis can allow an AI agent to identify novel hypotheses based on the past findings, and adapt strategies in real-time.

### *Agent short-term memory*

AI agents rely on short-term memory to process immediate information similar to working memory in humans. This buffer, context window, is finite and holds the agent's recent conversational history along with data from API responses, files to inform its actions (**Figure 4a**). Models like Gemini 2.5 pro and GPT-5 feature context windows of 1 million tokens over more, allowing them to process vast amounts of information simultaneously. This mechanism enables In-Context Learning (ICL), where the agent uses the provided context to generate responses without altering its underlying parameters[39,40].

### *Agent long-term memory*

To retain information permanently, agents utilize long-term memory, which exists in two forms, internal and external long-term memory. Internal Long-Term Memory is the parametric knowledge encoded within the neural network's weights during its initial training. This static knowledge base can be updated through several methods such as, periodically retraining the model on new, domain-specific to integrate new foundational knowledge (Continued Pretraining)[41], adapting the model for specific tasks, styles, or instruction-following capabilities (Fine-Tuning)[42], or combining the weights of multiple specialized models to create a single, more knowledgeable agent, which is useful when training data is proprietary (Model Merging)[43].

External Long-Term Memory (**Figure 4b**) provides a dynamic and persistent storage solution, most commonly implemented through Retrieval-Augmented Generation (RAG) systems. In a typical RAG architecture, external documents, scientific publications, or research data are embedded into numerical representations and stored in a vector database. This foundational approach has inspired more advanced techniques. For instance, Agentic RAG[44] employs an agent that can iteratively refine and reformulate search queries, improving retrieval accuracy. Similarly, GraphRAG[45], organizes knowledge into structured graphs and retrieves information by traversing entities and their relationships, rather than relying solely on text chunks. Julien *et. al.* show that GraphRAG can surface rare but important associations, such as identifying novel drug–gene–disease links for conditions like asthma and Alzheimer's disease outperforming classical RAG[46]. RAG enables an easily updatable, persistent memory without incurring the high cost of retraining the core model. At the same time, it typically relies on maintaining an in-house vector database, which introduces infrastructure overhead and requires ongoing manual updates.

### *Beyond the long-term memory*

As the field of drug discovery is rapidly evolving with new scientific literature, AI agents must extend their memory by actively interacting with data sources and other agents. Agents utilize a feature known as Tool Calling (**Figure 4c**) to interact with external databases and services via APIs which allows them to query specialized databases like ChEMBL for chemical data or PubMed for scientific literature, incorporating the results into their short-term memory. Model Context Protocol (MCP)[47] is an emerging protocol standard, developed by Anthropic, designed to create a universal interface between agents and external data sources.

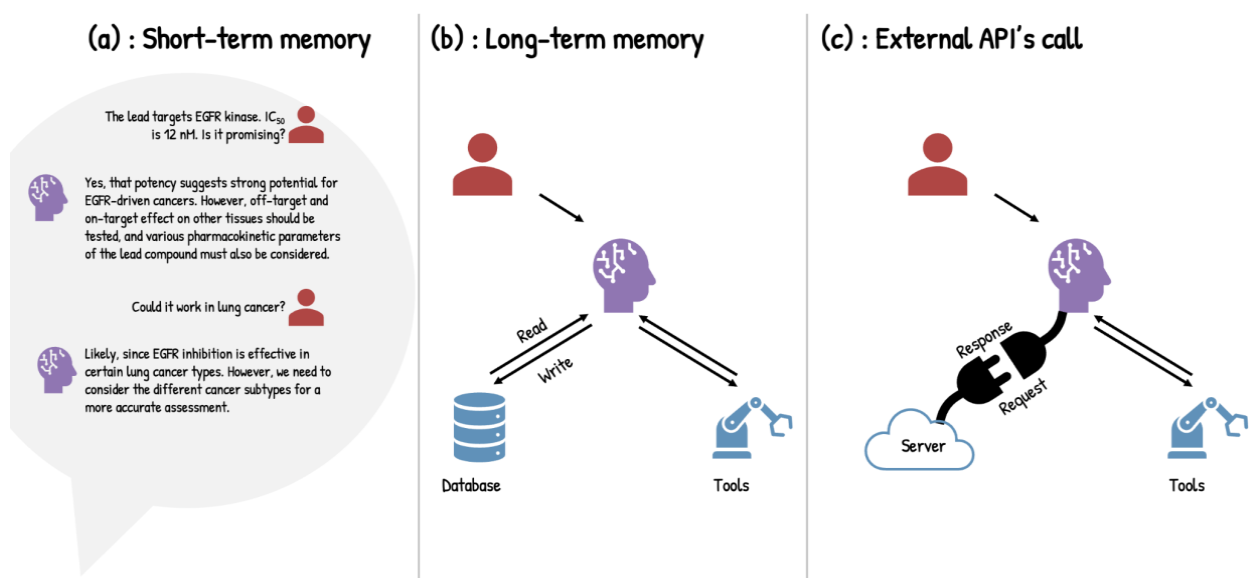

**Figure 4: Memory and External Integration in Agentic AI Systems.** (a) Short-Term Memory: The LLM maintains conversational context within the current interaction. (b) Long-Term Memory: The LLM can read from and write to an external database, allowing knowledge persistence across conversations. (c) External API Calls: The LLM interacts with third-party APIs to fetch real-time data.

# 3. Accelerating Drug Discovery with Agentic AI: Areas of Application

The first agentic AI systems have been in pilot in the drug discovery sector for around the past 12 months. Here, we present several case studies in which (multi)-agentic systems were developed to focus on specific functional areas of drug discovery and development. **Table S1** summarizes the domains, architectural designs of the agentic systems, key used tools, and notable outcomes for each case study.

## 3.1. Comprehensive Literature Analysis for molecular prioritization

Comprehensive literature analysis from large molecular datasets constitutes a tedious and time-consuming task in preclinical drug discovery. When medicinal chemists design multiple novel molecular candidates, assessing their likely properties requires searching relevant patents and publications for structurally similar analogs, extracting SAR data, cross-referencing conflicting measurements, and synthesizing insights. This process typically requires weeks depending on the scope and patent landscape, which then delays lead prioritization and synthesis decisions, particularly in early-stage discovery where rapid iteration is critical.

A multi-agent framework can transform this process by autonomously coordinating literature analysis tasks across distributed data sources (**Figure 5**). The system, developed by authors affiliated with Misogi Labs, employs a hierarchical supervisor pattern where an orchestrator agent coordinates specialized sub-agents, each with distinct domain expertise: (1) patent extraction agents retrieve relevant patents and extract SAR data using chemistry-aware parsing and RAG-enhanced retrieval, (2) literature retrieval agents query ADMET properties through semantic search across scientific publications, and (3) cross-reference agents identify conflicting data across sources using comparative analyses. A Model Context Protocol (MCP) standardizes database interactions, providing unified interfaces to ChEMBL, PubChem, and patent repositories. The RAG pipeline maintains citation provenance throughout the workflow, enabling each extracted data point to be traced back to its source document. A dual-memory architecture drives continuous improvement; short-term memory contains session-specific findings and intermediate results, while long-term memory retains scaffold-property patterns and query strategies, enabling recognition of previously analyzed chemical series across projects.

A typical workflow begins with uploading over a hundred novel molecular structures from a design library. The system autonomously: (1) performs similarity-based retrieval (Tanimoto similarity >0.7, substructure matching) to identify structurally related analogs from patent and literature databases, (2) extracts SAR data from associated documents for these analogs, (3) generates ADMET predictions for the novel compounds using proprietary models trained on combined internal and public datasets, covering properties including aqueous solubility, membrane permeability, CYP450 metabolic liability, and hERG cardiotoxicity, and (4) aggregates findings into structured reports with full citation provenance.

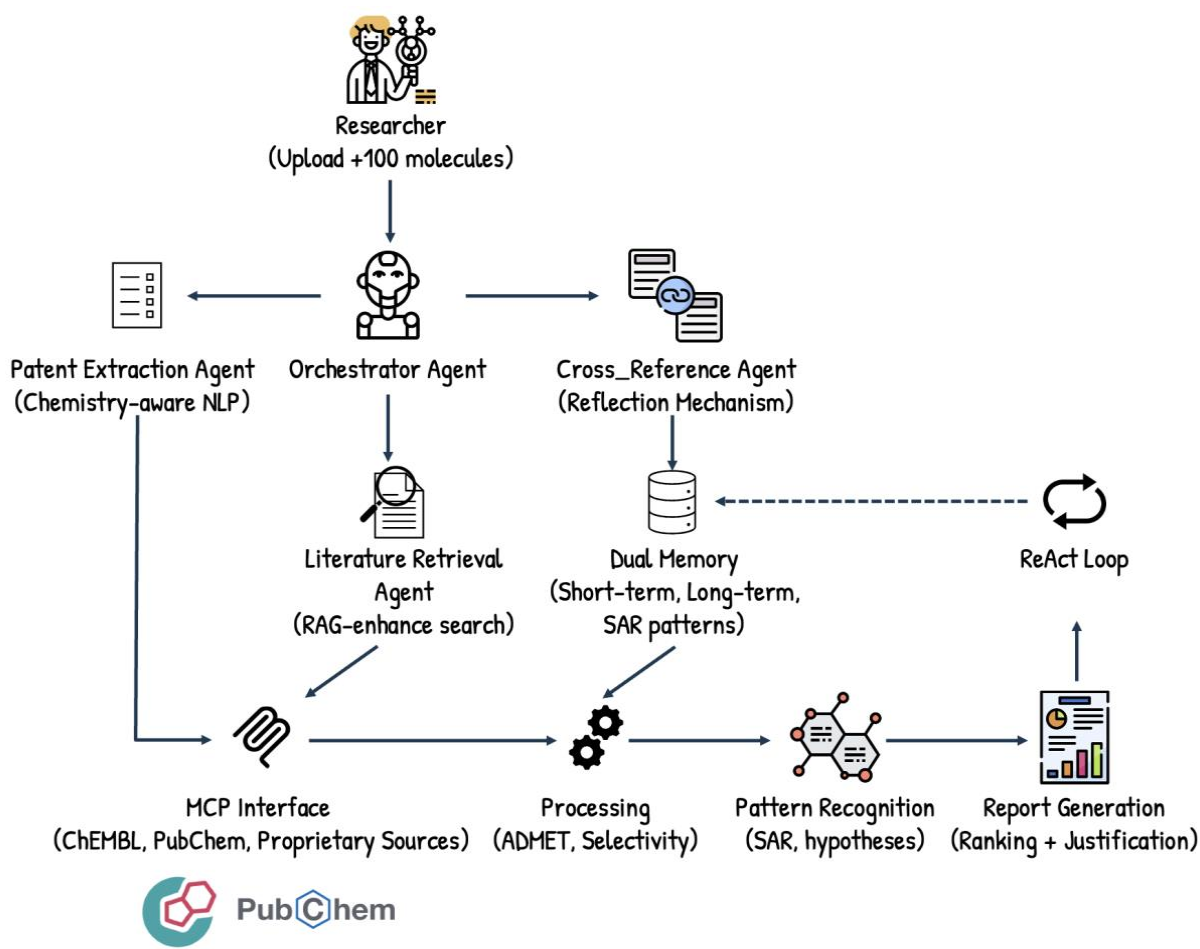

**Figure 5. Agentic system design for comprehensive literature analysis in connection to molecular structures / small molecules in early preclinical drug discovery.** The system's center is an orchestrator agent that harmonizes many different components, including Patent Extraction Agent, Cross-reference Agent, Literature Retrieval Agent and Dual Memory with molecular structure and data at the heart.

When sources report conflicting data for retrieved analogs (e.g., varying IC50 values for the same reference compound), cross-reference agents flag discrepancies and present all variants with experimental metadata (assay format, cell line, measurement conditions). Source metadata, including publication venue, citation frequency, and temporal recency, is provided to contextualize retrieved information. Critically, the RAG pipeline's document tracking ensures that all extracted data (IC50 value, ADMET measurement, or SAR annotation) can be traced back to its original patent or publication source, enabling us to verify claims and assess data reliability.

As a case study, authors affiliated with Misogi Labs initiated a BTK inhibitor discovery workflow where the orchestrator agent coordinated parallel operations across specialized sub-agents to assess acalabrutinib's selectivity profile for comparison with novel structural analogs. Acalabrutinib is a second-generation Bruton's tyrosine kinase inhibitor developed for the treatment of B-cell malignancies such as chronic lymphocytic leukemia and mantle cell lymphoma. It was designed to achieve improved selectivity and reduced off-target kinase inhibition compared to first-generation BTK inhibitors like ibrutinib. Patent analysis agents retrieved relevant drug patents through similarity-based search on Morgan fingerprints (radius=2, n-bits=2048, Tanimoto Similarity >0.7), literature retrieval agents queried scientific databases for kinase selectivity data (**Table 2**) and ADMET properties, and cross-reference agents identified conflicting measurements across sources (**Table 3**).

Table 2. Multi-Source IC$_{50}$ Data Extraction - Acalabrutinib and Multiple Kinase Targets

| Target Kinase | Biochemical IC$_{50}$ (nM) | Cellular IC$_{50}$ (nM) | Inter-assay Range | Assay Type (Cellular) |
|---|---|---|---|---|
| BTK | 3.0–5.1 (consensus: 5.1) | <10 | Low variability | Z'-LYTE, IMAP |
| TEC | 9.7±2.6 (Hopper 2018)<br>37–126 (other studies) | >1000 | 13-fold variability | Platelet phosphorylation |
| BMX | 46–598 (consensus: 58) | N/A | 10-fold variability | n.a. |
| EGFR | >1000–3513 | >10,000 | Consistent | A431 EGF-induced |
| ERBB2 | >1000 | >1000 | Consistent | Z'-LYTE |
| ERBB4 | 16–140 | N/A | 9-fold variability | n.a. |
| ITK | >1000–30,000 | >1000 | Low above threshold | T-cell assays |
| JAK3 | >1000 | N/A | n.a. | n.a. |

Table 3. Conflict data detection and justification

| Conflict ID | Target | Biochemical IC$_{50}$ (nM) | Cellular IC$_{50}$ (nM) | Fold Difference | Root Cause | Clinical Impact |
|---|---|---|---|---|---|---|
| CONFLICT-1 | TEC | 9.7±2.6 (Hopper 2018) 37–126 (other assays) | >1000 (platelet assay) | >100-fold (biochem→cellular) | Biochemical IC$_{50}$ shows only 3-4× selectivity vs BTK Cellular assays reveal NO functional TEC inhibition | CRITICAL: Despite low biochemical selectivity, cellular data confirms no platelet TEC inhibition, demonstrates disconnect between binding and function |
| CONFLICT-2 | TEC | 13-fold inter-assay variability (9.7–126 nM) | Consistent >1000 nM | Variable | Different biochemical assay formats (mobility shift, Z'-LYTE, kinetic) | MAJOR: Highlights assay-dependent results; kinetic studies most rigorous |
| CONFLICT-3 | BTK | 3.0 vs 5.1 nM | <10 nM (both) | 1.7-fold | IMAP vs Z'-LYTE assay format | MINOR: cellular activity consistent |

The system delivers a comprehensive report with structured molecular entity tables, extracted SAR data with patent citations, conflict flags with source metadata, ADMET prediction profiles, and complete citation chains to original documents. Overall, this agentic workflow enables us to verify each claim against source materials and assess data reliability based on experimental conditions, allowing them to focus on interpretation and synthesis prioritization rather than manual literature aggregation.

## 3.2. *In Silico* Toxicity Predictions

The pharmaceutical and chemical industry faces several bottlenecks in toxicological assessments. A chemistry-specialist Agentic AI can greatly benefit chemical toxicity prediction, combining approaches in predictive toxicology modeling and cheminformatics for safer chemical design.

For *in silico* toxicology predictions, the authors affiliated with Human Chemical Co. built an agentic AI system using the ReAct architecture that conducts a series of observations, decisions, and tasks. It consists of a suite of predictive and cheminformatics tools and can summarize and contextualize predicted values with a comprehensive, structured overview of existing scientific literature, regulatory assessments, functional uses, and other relevant information. After the completion of each cycle, users can follow up with questions or further tasks, based on which the cycle continues to run and returns further results. This human-in-the-loop architecture increases research productivity by an order of magnitude while continuing to depend on the human expertise necessary for chemical safety assessment.

As an example, authors affiliated with Human Chemical Co. used this Agentic AI system to assist in the safety assessment of the endocrine disruption risk of the chemical compound Cashmeran. Cashmeran is a synthetic musk compound developed in 1969, used as a fragrance ingredient, chiefly in perfumes. It belongs to a class of synthetic musk compounds known as polycyclic musks, named after their shared structural feature of multiple fused or bridged carbon rings. After querying the Agentic AI system on Cashmeran using its CAS number (33704-61-9), the system automatically initiated a series of toxicological property predictions using proprietary models. Finally, the system retrieved, synthesized, and returned a comprehensive report regarding the known properties of cashmeran, with sources including the scientific literature and authoritative regulatory statements.

The system indicated an endocrine disruption hazard for the compound Casherman, which was also shared across the class of polycyclic musks. To move from hazard to risk assessment, the Agentic AI System retrieved two predicted metabolites from the open-source BioTransformer 3.0 tool. It then re-ran in silico prediction of endocrine disruption hazard for these metabolites and found a rapid reduction in hazard from the Cashmeran parent compound. The system repeated this endocrine hazard prediction process for the metabolites of other common polycyclic musks, revealing Cashmeran's metabolism to be relatively rapid and its hazard profile to be highly favorable (**Figure 6**).

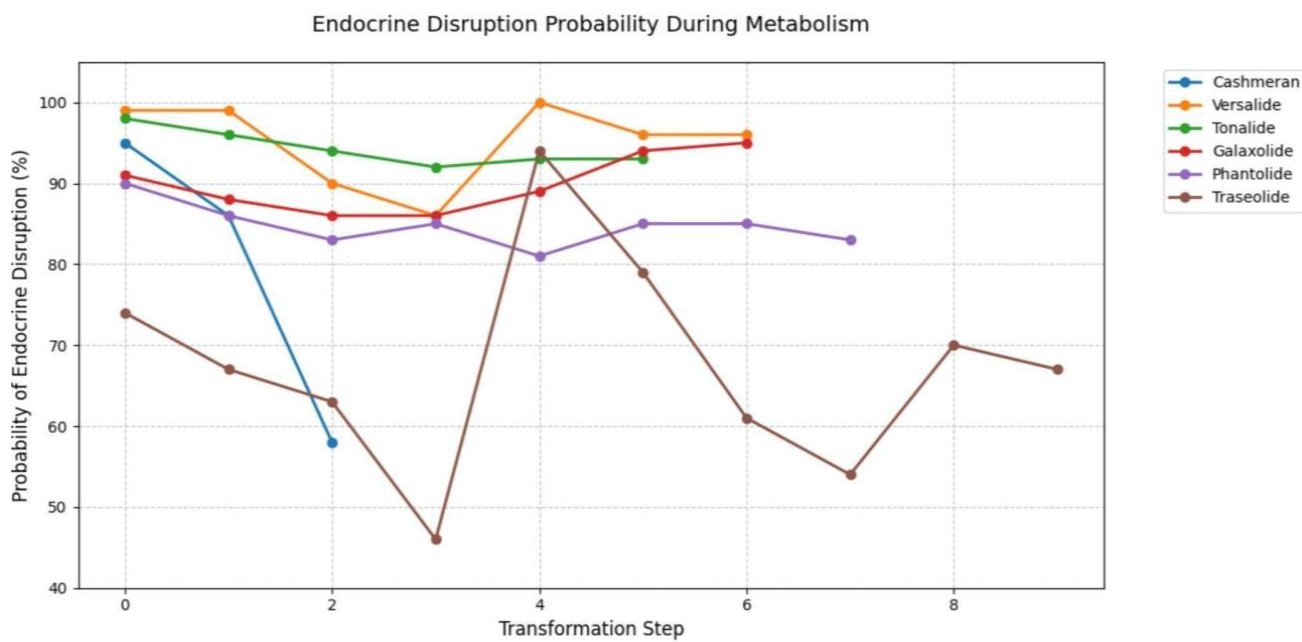

**Figure 6.** Endocrine disruption probability of six chemicals during the metabolism steps.

The agentic AI system also generated some predicted Cashmeran metabolites using its cheminformatics tool, calling and repeating the hazard prediction process for these metabolites. This additional analytic process replicated the findings of the initial metabolite hazard assessment, with a downward trend in predicted endocrine hazard observed over the metabolic process. The results are summarized in **Table 4** below, which the Agentic AI System generated.

**Table 4.** Hazard prediction results of Cashmeran metabolites

| Metabolite | Structure Description | Structure | Endocrine Activity Probability | Confidence Score | Comparison to Parent |
|---|---|---|---|---|---|
| Parent Cashmeran | Original compound | | 73.7% | 1.0 | - |
| Aromatic Hydroxylated | Hydroxyl group added to aromatic ring | | 61.6% | 0.5 | ↓ 12.1% |
| Aliphatic Hydroxylated | Hydroxyl group added to methyl side chain | | 60.7% | 0.75 | ↓ 13.0% |
| Ketone Reduced | Ketone reduced to alcohol | | 61.7% | 0.875 | ↓ 12.0% |
| Demethylated | One methyl group removed | 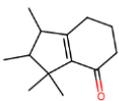 | 70.2% | 1.0 | ↓ 3.5% |
| Carboxylic Acid | Oxidation to carboxylic acid | 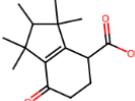 | 45.6% | 0.125 | ↓ 28.1% |

Simultaneously, the agentic AI system conducted a comprehensive scientific literature review for Cashmeran, and from this assessed its endocrine disruption risk to be low. The system provided several justifications for its assessment, including: published *in vitro* high-throughput screening data from ToxCast, which found found Cashmeran to be inactive in estrogen-related assays at non-cytotoxic concentrations[48]; a lack of adverse endocrine-related effects reported in a 90-day repeated gavage dosing study of Cashmeran (ibid.); a classification of "not toxic" by the Australian Industrial Chemicals Introduction Scheme; and an *in vitro* observation of the rapidity of Cashmeran metabolism relative to eight other fragrance chemicals, with a further observation of notably faster metabolic rates in humans compared to trout (half-life <1 hour in humans versus <1 day in trout)[49]. Also cited as related supporting evidence was a relatively high NOAEL of 1875 mg/kg/day (oral, rat) established in a reproductive toxicity assay for Cashmeran[50] and an ECHA report of "no adverse effect observed" regarding toxicity to reproduction for Cashmeran[51].

When queried to assess its *in silico* predictions of Cashmeran's endocrine disruption hazard in light of the published literature, the Agentic AI System found significant agreement between its predicted and retrieved results. Using structure-activity relationship analysis, the system conjectured that its final predicted metabolite for Cashmeran (a carboxylic acid) demonstrated reduced probability of endocrine disruption due to its relatively high polarity and consequent poor membrane penetration and reduced receptor affinity. It also noted that, while Cashmeran has sometimes demonstrated weak estrogenic activity *in vitro*, such assays assume direct exposure of the parent compound to animal estrogen receptors without taking metabolism into account, which could explain for the disconnect between occasional *in vitro* hits and the absence of observed *in vivo* endocrine risk.

These results demonstrate the use of Agentic AI systems in chemical safety assessment, especially when real-world exposure scenarios call for complex, multi-step toxicological analyses combining computational predictions with experimental data. Such systems are expected to have application not only in drug discovery, but in cosmetic development, industrial chemistry, materials science, and environmental toxicology.

## 3.3. Automating Protocol Design and Execution

Designing and validating molecular assays like quantitative polymerase chain reaction (qPCR) remains a slow and expertise-intensive process. Even routine assays require us to spend days reviewing literature, comparing methods, identifying validated reagents and primers, and then translating the protocol into executable steps for liquid-handling systems. Regulatory alignment adds additional complexity, demanding traceable documentation and adherence to standards such as MIQE or FDA guidance for assay validation. Such validation activities can take up to several months, making up a meaningful component of many drug development timelines. In traditional workflows, this process involves multiple disconnected stages, including literature review and protocol drafting by experimentalists, and coding by automation engineers. The lack of integration between these steps increases coordination time, cost, and variability across laboratories.

A generic agentic system can address these inefficiencies by coordinating the entire assay-development workflow, including reading literature, reasoning about experimental design, generating protocols, and producing executable lab instructions. Equipped with perception tools for literature retrieval, reasoning modules for comparative method assessment, and action tools for robotic code generation, agents can autonomously design and refine assays. For example, an agent can retrieve and evaluate published qPCR methods, identify those with highest sensitivity and specificity, generate a MIQE-aligned protocol, and translate it into executable code for automation platforms such as Opentrons. By connecting these steps through a closed-loop reasoning process, the system transforms static knowledge into executable workflows, reducing human overhead while improving reproducibility and traceability.

As an example, authors affiliated with Potato developed a multi-agent system, *Tater*, that integrates retrieval-augmented generation (RAG) with specialized scientific tools spanning planning, execution, and iteration. A scientist provided the following task: "Derive an automated protocol on an Opentrons to perform qPCR quantification of AAV. This is for pre-clinical grade material but the assay should be able to be developed into a clinical release test. Search the literature for appropriate methods, identify those with the best sensitivity and specificity that meet FDA requirements, and write a detailed protocol including what data analysis needs to be done."

Within two hours, the system generated a curated literature collection and citation map of qPCR methods used for AAV quantification. This was followed with a comparative table assessing assay parameters (sensitivity, specificity, MIQE alignment, and regulatory suitability). A complete qPCR protocol with step-by-step operational details and required controls was then provided (**Figure 7**). The agent further used an opentrons automation script to translate the method into executable code for a liquid handler. Finally, it presented a structured report detailing assay scope, analytical criteria, automation considerations, and statistical analysis templates. Overall, the output demonstrated reasoning capabilities that extended beyond static text summarization. The agent autonomously integrated information from multiple sources, synthesized a coherent and validated experimental workflow, and produced executable code for laboratory automation systems. This end-to-end process exemplifies how agentic systems can translate distributed scientific knowledge into operational protocols suitable for immediate execution on lab hardware.

Compared to the manual process of designing and validating an AAV qPCR assay, Tater reduced total setup time and human effort by more than two orders of magnitude across the workflow (**Table 5**).

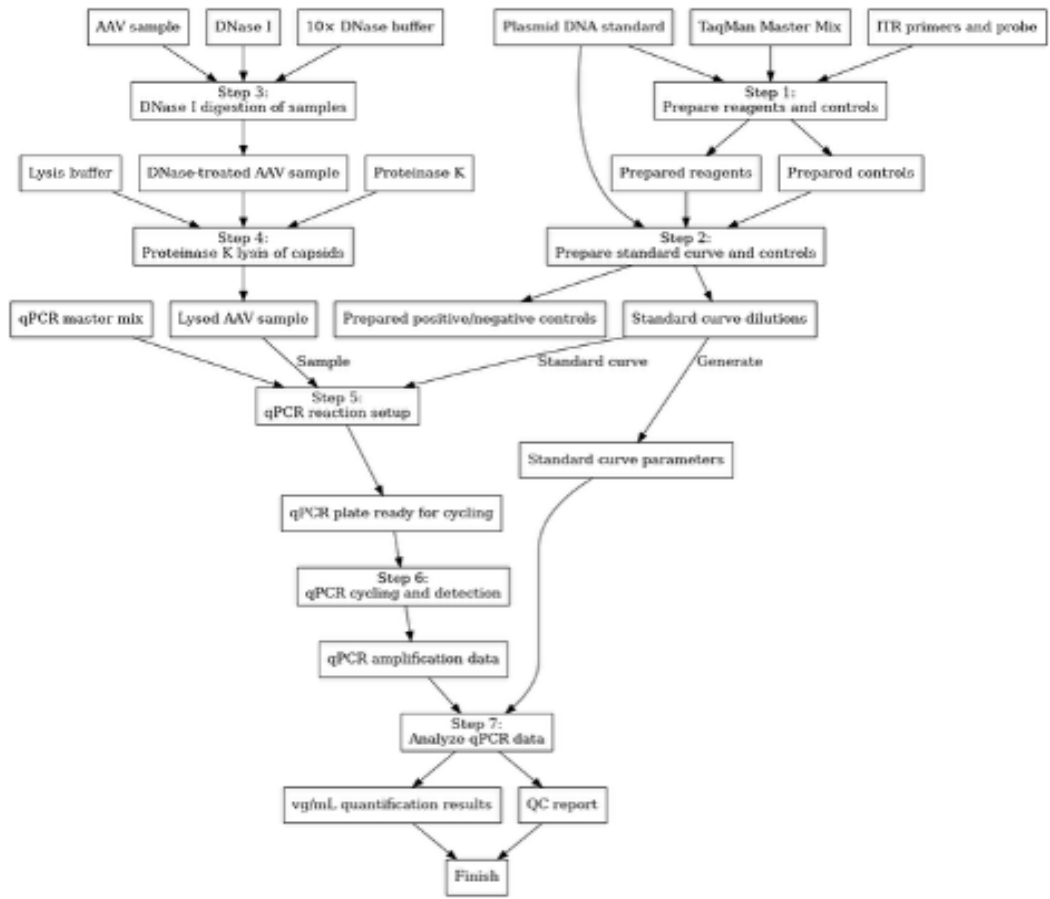

**Figure 7. Stepwise workflow of the qPCR protocol generated by the agent.** The figure depicts the complicated and comprehensive experimental sequence produced by the Protocol Generator, and then run by Tater.

**Table 5. Automating qPCR assay development using Potato's Tater agent.** (a) Manual qPCR design process requiring weeks of literature review, manual protocol drafting, and scripting by automation engineers. (b) Tater's integrated agentic workflow linking literature retrieval, reasoning, protocol generation, and automation script output. (c) Example outputs include comparative methods table, MIQE-aligned protocol, and Python-based Opentrons code, and (d) Quantitative comparison demonstrating compression of the four-month workflow to under two hours.

| Workflow Stage | Manual (Typical lab estimate) | With Tater | Δ Improvement (Based on min manual estimate) |
|---|---|---|---|

| Literature review & method comparison | 5 - 14 days | 13 min | ≈550× faster |
|---|---|---|---|
| Protocol drafting & MIQE validation | 6 - 10 hours | 8 min | ≈45× faster |
| Translating protocol to script | 3 weeks - 3 months | 55 min | ≈260× faster |
| Total end-to-end design cycle | 1 - 4 months | 1 hr 39 min | ≈400× faster |

When accounting for downstream verification and testing by laboratory experimentalists, the efficiency gains remain substantial. Tater's outputs, including literature syntheses, protocol drafts, and automation scripts, require human review to confirm reagent compatibility, instrument calibration, and compliance with local quality standards. These verification steps typically add hours rather than weeks to the overall process. In effect, the agent reduces the human effort required to arrive at a validated, automation-ready assay by several orders of magnitude while maintaining full traceability and auditability of its reasoning chain. Nonetheless, automated protocols must still undergo empirical validation to confirm experimental performance and regulatory compliance, ensuring that speed gains do not compromise assay reliability. Rather than replacing scientific judgment, the multi-agent system enables us to shift from manual document assembly to higher-value experimental reasoning, optimization, and validation.

By autonomously reasoning from literature through to executable automation code, an agentic AI system compresses what used to require multidisciplinary coordination into a single, auditable workflow. A process that traditionally spans months of cross-functional effort was completed in under two hours, representing a >400× reduction in cycle time. This result establishes a quantitative benchmark for next-generation autonomous scientific systems, demonstrating how automation can augment human expertise while preserving scientific integrity and oversight.

### 3.4. Accelerating Drug Discovery with Virtual Scientists

Drug discovery workflows are often fragmented across biology, chemistry, and clinical domains. We must integrate heterogeneous data, tools, and models, but conventional AI systems often lose experimental context, treat tasks as isolated, and make it hard to prioritize experiments while maintaining transparency, reproducibility, and scalability.

One way to accelerate drug discovery is through an infrastructure-focused platform powered by Virtual Scientists which delivers a multi-agentic AI system, supporting us across all the activities in drug discovery. The Virtual Scientists, here implemented by authors affiliated with Kiin Bio, integrated over 100 data, tools, and AI models across biology, chemistry, and clinical domains, from API-based tools to large-scale bioinformatics pipelines and GPU-based models, enabling it to propose, evaluate, and prioritise experiments, both *in silico* and lab-based, while ensuring transparency, reproducibility, and scalability. Virtual Scientists are defined as specialised AI agents each designed to take on a distinct scientific role (**Figure 8**), including agents that (1) conducts literature reviews, explores biological data, and identifies relevant datasets, cell lines, and competitor information, (2) runs large-scale pipelines, performs multi-omics analyses such as transcriptomics and proteomics, and carries out tasks like variant analysis, GWAS and polygenic risk scoring, and (3) focuses on molecular design and screening, including virtual screening, protein folding and docking, property prediction and novelty checking Each agent can work autonomously but also contributes to a shared organisational memory, enabling coordinated decision making across disciplines and teams.

A core innovation of a Virtual Scientist system lies in its ability to capture and reason over experimental context. AI platforms often lose this context, treating tasks as isolated problems. This can be addressed by embedding context into every stage of orchestration, ensuring that Virtual Scientists understand not just the question posed, but also how it connects to prior evidence and broader project goals. This context awareness enables more accurate prioritization of experiments, reduces redundancy, and increases reproducibility, ultimately accelerating the path to meaningful scientific insight.

As a case study, authors affiliated with Kiin Bio have been supporting a partner to develop a new pre-clinical drug discovery program for Idiopathic Pulmonary Fibrosis (IPF). The objective here was to identify the market opportunity, prioritize novel therapeutic targets, then generate and rank potential small molecule hits. The Virtual Scientists successfully orchestrated this non-linear workflow (**Figure 8**), integrating multiple Virtual Scientist agents to execute and reason across public omics data, scientific literature, RNA-seq analytics, structural biology models, and generative chemistry models. Examples of outputs from the various stages of this automated workflow included comprehensive disease landscape review, ranked set of disease relevant omics datasets as well as an initial set of small molecules with predicted binding affinity to target of interest. The Kiin Bio Virtual Scientists are guided by the experiment-level context defined by the users at the start of the process. This experiment-level context grows with the experiment and is used directly to provide next activity suggestions to the user. The process could then be executed end-to-end in under two hours (depending on dataset size), while it usually takes 2-3 weeks to conduct such activities in real life. Additionally, the system is executed with the flexibility to rerun all or part of the workflow as new experimental data becomes available. Overall, processes such as molecular docking and generative chemistry were substantially streamlined through AI-agent orchestration. Nonetheless, further advancement is needed in the integration of heterogeneous omics and structural datasets, where inconsistencies in normalization and annotation standards can introduce analytical variability. In parallel, operational tasks such as contract research organization (CRO) selection, which traditionally require multiple rounds of communication over several months, were executed with markedly improved efficiency through automated coordination and decision-support agents.

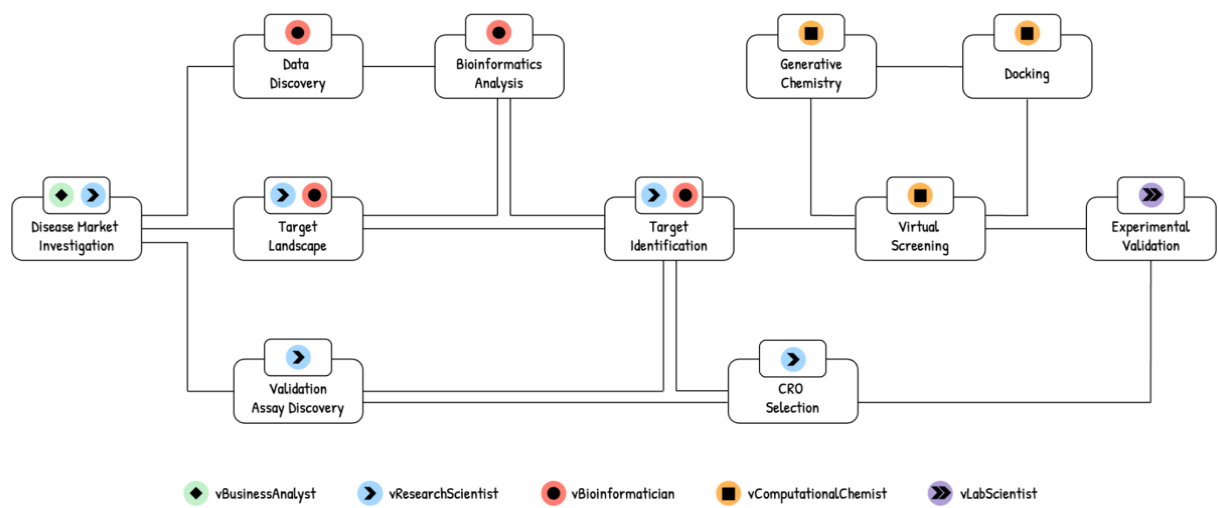

**Figure 8. Drug discovery workflow enabled by Kiin Bio's Virtual Scientist Platform**. Each block represents a distinct research plan within a larger experimental workflow. Plans are tagged with the Virtual Scientist agent responsible, illustrating how specialized agents collaborate to execute and coordinate complex end-to-end discovery processes.

## 3.5. Drug Repurposing for Rare Diseases

Rare diseases are associated with inherently small patient populations and the scarcity of understanding of the data. In the case of drug repurposing, traditional discovery relies on cross-referencing heterogeneous datasets to uncover existing drugs on the market that could have new therapeutic applications[52]. However, cross-referencing manually is time-consuming, resource-intensive, and prone to missing some connections hidden across biology and chemistry datasets. Knowledge graphs have emerged as powerful tools for integrating biomedical data in rare disease drug repurposing. *Yuryev et. al.* demonstrated that such graphs can uncover drug-disease associations from both open and restricted-access literature[53]. Yet, despite the promise, knowledge graphs remain largely static resources, limited to surfacing associations without the ability to actively reason across heterogeneous data sources, adapt to new evidence in real time, or coordinate the multi-step processes required for drug repurposing.

To address this, an agentic MCP-driven system was used to accelerate repurposing using a supervisor architecture as the core design (**Figure 9**). A supervisor coordinated a suite of specialized agents, each with a distinct role in the discovery pipeline. A disease agent retrieved associated genes and variants from resources such as Ensembl and OpenTargets, while a pathway agent mapped relevant biological processes using Reactome and KEGG. In parallel, a protein agent gathered structural information from AlphaFold and the Protein Data Bank (PDB), and a compound agent searched databases like ChEMBL and PubChem to identify existing molecules that target those proteins. Finally, a safety agent evaluated toxicity and ADMET profiles to filter out unpromising candidates, ensuring that only the most promising molecules are returned through the pipeline.

As a case study, the authors associated with Augmented Nature used this system with the following query: *"Find repurposing opportunities for spinal muscular atrophy (SMA)"*. The orchestrator initiated multiple agents simultaneously (**Figure 9**). The disease agent identified key genes linked to SMA, such as SMN1 and SMN2, while the pathway agent mapped the biological mechanisms they influence. Structural data for relevant proteins was then retrieved, and compound agents searched for ligands that interact with these targets. The safety agent applied an additional layer of filtering by prioritizing candidates with favorable ADMET properties. This fully automated, parallelized approach dramatically reduced the manual burden while ensuring comprehensive data coverage.

Overall, an agentic MCP-driven system generated a shortlist of potential repurposable ligands within a few hours, compared to the weeks required by manual human-centric approaches. The coverage of such a search is broad, and it integrates genomic, pathway, structural, and chemical evidence into a single pipeline similarly to[54]. From a practicality standpoint, the system tends to surface drugs that are already approved or in clinical development, possibly lowering regulatory and financial barriers to further testing. Most importantly, the approach is scalable: the same infrastructure can be readily applied to any rare disease with minimal setup, making it a powerful tool for speeding up the early discovery phase.

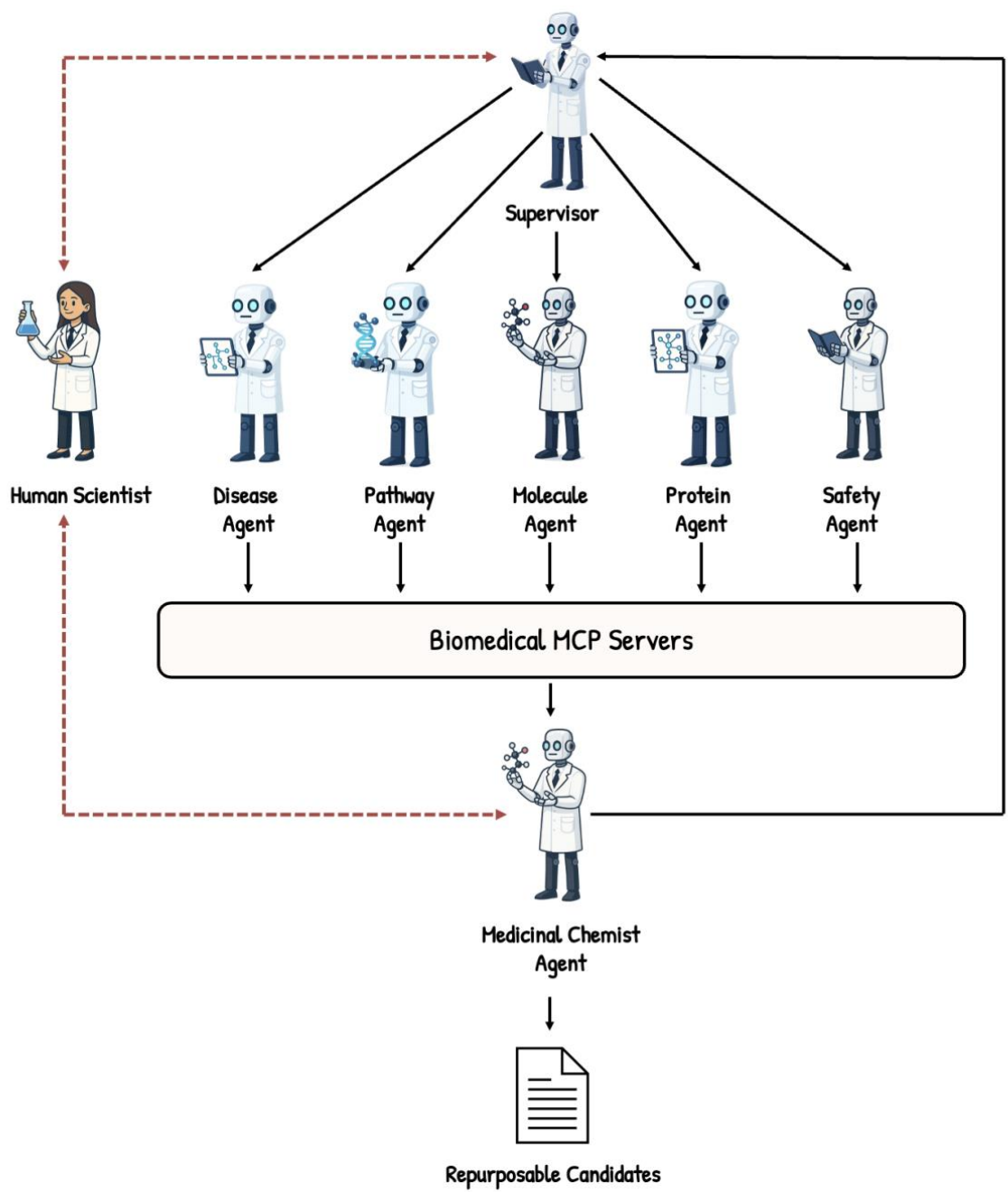

**Figure 9. Agentic System to accelerate drug repurposing for rare diseases.** This system uses the supervisor multi-agent architectures, with 5 sub-agents, each is responsible to work with one specific domain, including Disease Agent, Pathway Agent, Molecule Agent, Protein Agent, and Safety Agent. At the termination, there is one Medicinal Chemist Agent that can validate the results. Humans can interact with Supervisor and Medicinal Chemist agents.

## 3.6. Automating small molecule synthesis

Synthesis of candidate compounds often constitute a rate-limiting step in the small molecule discovery process[55]. In some cases, the synthesis process can take up to six months and cost tens of thousands of dollars or even more. Automation of this process presents a big opportunity to shorten iteration cycles and decrease drug development costs. The problem of small molecule synthesis automation is two-fold[56]. The investigation of the compound's synthetic route is essential, given that almost any compound encountered during drug development will not have a published method of synthesis. To put this into perspective, there are $10^{60}$ drug-like compounds, but only hundreds of millions are present in such databases as SciFinder or Reaxys. One would also need to find a way to recover from failed experiments; perhaps the reaction mixture still can be saved by adding another reagent, a change of solvent, an increase in temperature, or another synthetic route can be implemented. The execution of the synthesis reaction needs to be automated. Chemistry raises several challenges here: harsh conditions, inert atmosphere, corrosive compounds, and solvents that can dissolve parts of the apparatus.

In the paradigm of AI agents, these systems can help with both aspects of determining the path to synthesis and automating execution. The intelligence layer of synthesis work is an obvious application. Standard retrosynthesis tools and databases become tools for agents where scientific papers can be retrieved and analyzed. The ability of models to work with chemical data and "understand" chemistry is critical for this task, and thus robust benchmarks must be used to track performance over time. Execution in the physical world is a more challenging but necessary step to achieve full autonomy in organic synthesis. Several works attempted tight integration between experiment execution hardware and LLM agents[25,26,57,58]. Following on from this work, an agentic platform was developed by authors affiliation to onepot.ai which can execute organic chemistry experiments in the physical world via action tools. On the hardware side, the system consisted of a tube picker, a tube decapper, a liquid handler, a plate sealer, and an LC/MS system for analysis and purification. On the virtual side, the system developed has access to all the internal knowledge about the experiments performed; this helps models rely on the experimental results in addition to the broad knowledge corpus it was exposed to during training. These two sources of information are crucial to make a complete version of an "AI scientist", a system that can not only generate novel ideas and hypotheses but also validate them experimentally and learn from the results over the long horizon. In addition, a suite of tools streamlines the research operation.

The authors affiliated with onepot.ai used this system, which includes a retrosynthesis engine for planning the experiments, web search, code execution for robust analysis, and a literature search to augment the process in designing the experiments. They designed an "AI organic chemist" capable of executing experiments and modifying protocols as needed in a loop (**Figure 10**). The system was used for organic synthesis for commercial and in research settings. The integration of the experimental setups with agentic systems makes it closer to a fully closed-loop discovery. Advancements in LLM, such as reasoning, custom tools, and agentic capabilities, significantly benefit the intelligence that goes into the experiment planning, analysis, and novel hypotheses generation. As a result, this system enabled commercial operation of an automated small molecule synthesis facility covering 7 distinct reactions (with 2 to 5 protocols per reaction, and constantly expanding, having average synthesis success rate ranging from 50% to 88%, depending on the reaction). Overall, throughput of the system allowed for synthesis of tens of compounds per day.

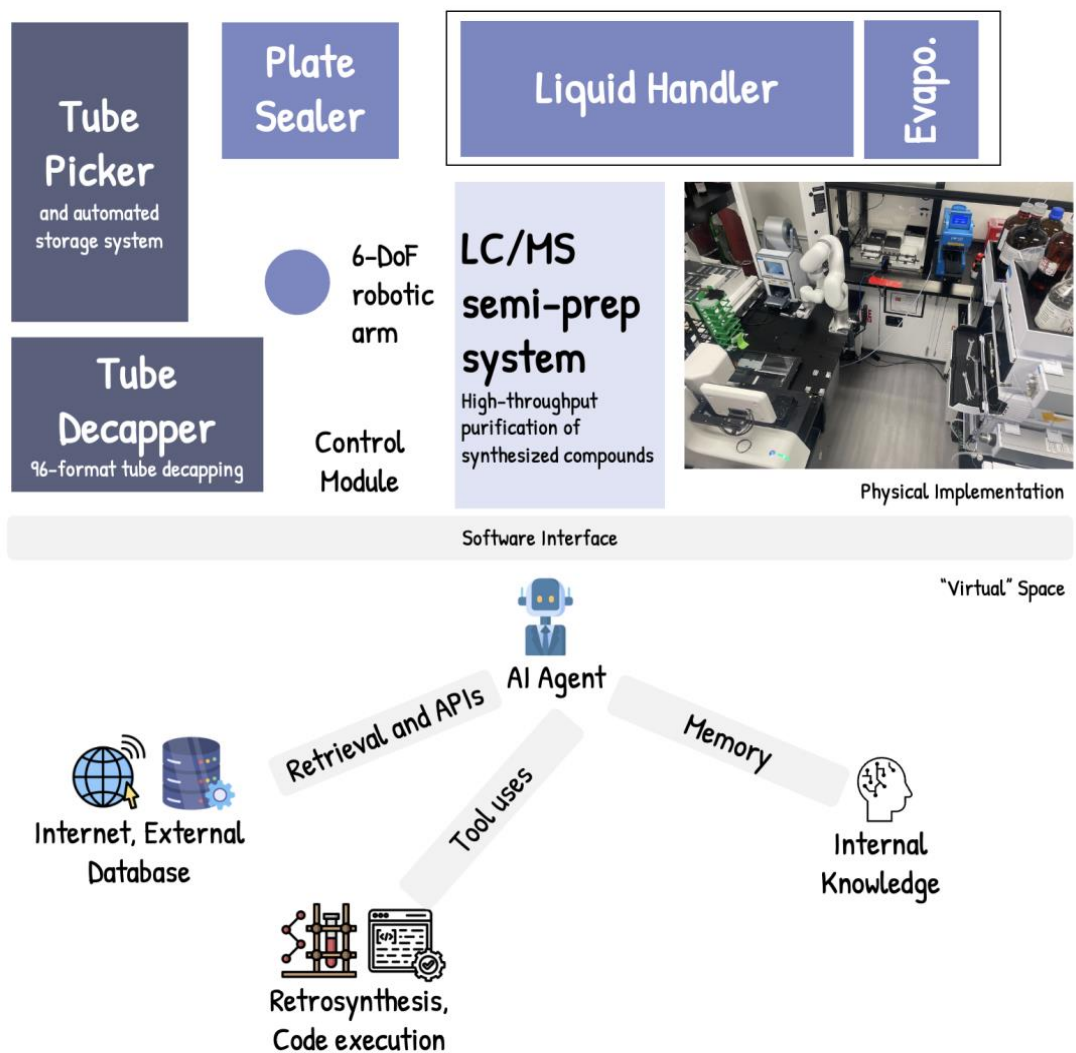

**Figure 10. Automating small molecule systems with hardware automation and AI.** An agentic system for organic chemistry experimentation is split into physical implementation and virtual space.

## 3.7. Augmenting drug discovery workflows with better search via focal graphs

Large-scale biomedical datasets present several challenges for data mining; they are often complex, diverse, noisy, sparse, siloed, and poorly annotated. Graph-based approaches, such as PageRank (used in internet search) and crowd sourcing, are well-suited to retrieve information from datasets such as these, making them an attractive foundation for AI agents. Internet searches, which focus on interlinked text-based resources, are already used routinely in RAG-based applications. In this case study, focal graph approaches have been implemented to generalize this to a more abundant non-textual, minimally processed data[59].

While comprehensive knowledge graphs provide a mode of data representation amenable to analysis by network algorithms, their direct analysis can be computationally expensive and their visualization unwieldy. In advanced graph analysis and information retrieval, focal graphs are portions of a larger graph that have been extracted because they hold information pertinent to a particular query. By concentrating on the most relevant data points, they simplify the exploration of large datasets, preserve meaningful relationships and context, and decrease the computational resources required for analysis. (**Figure 11**). When diverse query types, including gene expression profiles, chemical structures, lists of cell lines, etc, are used, entities most strongly associated with the queries, and therefore central to the focal graph, can be surfaced, and the specific data connections behind them can be exposed, making the results both concise and transparent.

Individual focal graphs searches have been used to generate data-driven hypotheses across multiple applications, including uncovering putative mechanisms driving cellular phenotypes in a cell painting data set[60], discovering new molecular targets for compounds based on their chemical structure, and identifying potential off-target toxicity mechanisms based on chemical structure and transcriptomics data[61] (unpublished results). As a case study, authors affiliated with Plex Research used focal graph-enabled LLMs to conduct complex, extended, *in silico* research programs that produce novel, verifiable, data-driven hypotheses. A focal graph-enabled AI agent, prompted with, "Please plan and execute a research program to identify a novel oncology target in the Wnt pathway." proceeded to list well-known members of the Wnt pathway, identified RNA seq profiles where these pathway members were perturbed, and then ran focal graph searches to identify other, potentially novel genes, which when perturbed, give rise to similar gene expression profiles. This systematic phenocopier approach led to the identification of several potentially novel oncology targets in the Wnt pathway, including the eIF2 complex[59]. Overall, the workflow shows that *in silico* efforts using AI agents can be augmented with ability to search data across essentially any therapeutic area, treatment modality, data type, or pipeline stage.

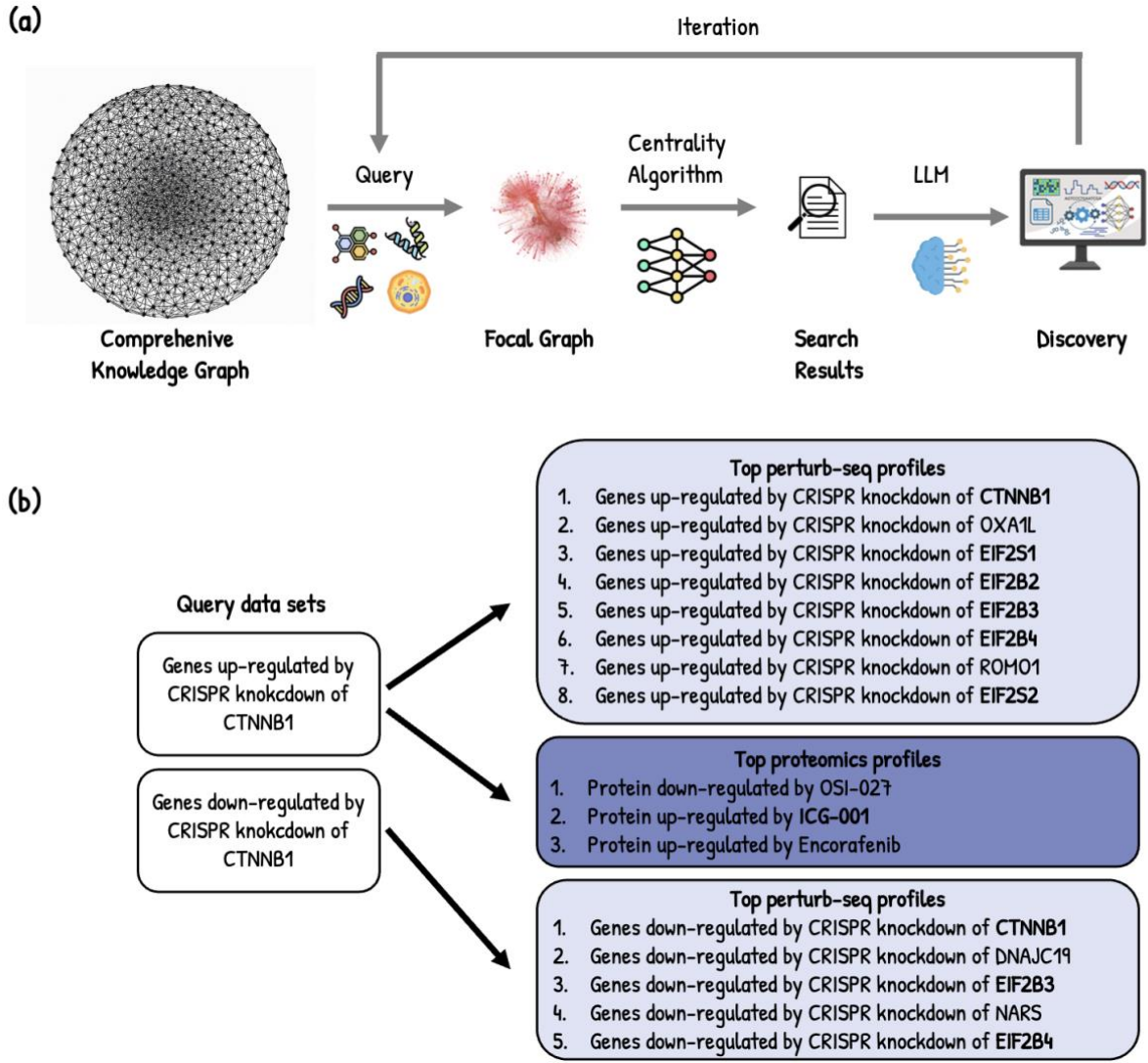

**Figure 11. Workflow that uses focal graph as data source for LLM's reasoning.** (a) The focal graph is extracted from a comprehensive knowledge graph, which then applies a centrality algorithm to provide informational data that supports LLM reasoning. (b) Discovery of novel components in the β-catenin pathway by focal graph analyses of genes up- or down-regulated after CRISPR knockdown of β-catenin[53].

## 3.8. Discovery-to-Deal decisions

Early-stage biopharmaceutical asset discovery is hindered by manual, fragmented processes where critical data about drug assets is dispersed across multilingual and unstructured sources. Conventional tools often fail to connect and find assets due to transliteration and alias drift, while the integration of scientific merit with commercial, operational, and regulatory strategy remains an inconsistent, bloated task. A multi-agent system can automate the synthesis of these disparate domains, performing end-to-end asset discovery, evaluation, and strategic packaging.

In this case study, authors affiliated with Convexia Bio developed an agentic workflow for small molecule asset search and evaluation (**Figure 12**). The system incorporates a feedback loop that enables continuous refinement through five core modules: (1) Asset Discovery that sources and retrieves data from multiple sources and creates a query-specific knowledge graph that integrates normalized, provenance-linked evidence, (2) Scientific Evaluation that runs computational simulations to predict drug properties and assess technical merit, (3) Market Analysis that generates market landscape, regulatory pathway, and intellectual property scenarios, (4) Clinical Assessment which scores clinical and operational risk through predictive simulations based on historical trial data, and (5) Business Development which, calculates the best path forward using analog deals and scenario modeling..

Users provide queries specifying parameters such as target, geography, modality, development stage, and delivery route. The system retrieves and validates data from structured and unstructured sources, storing all evidence in a centralized knowledge graph that downstream modules can access without redundant data collection. The technical evaluation combines physics-based methods (molecular docking, molecular dynamics, free energy calculations) with machine learning models trained on curated datasets. When predictions from different methods disagree, the system flags uncertainty and may suggest additional experimental validation rather than forcing a conclusion. The clinical module leverages historical trial outcomes to predict risks related to contract research organization reliability, manufacturing readiness, site activation timelines, enrollment patterns, and protocol complexity. Real-world data from regulatory filings, vendor disclosures, and clinical registries continuously update these predictions. Market and business development analyses combine computational tools with data retrieval from claims databases, analyst reports, regulatory filings, health technology assessments, and pricing databases. Financial projections incorporate patient population estimates, pricing, market penetration, and costs. Partner matching uses historical deal patterns to identify suitable buyers and generate tailored partnership proposals. All analyses, assumptions, and data sources are documented in the knowledge graph for transparency and reuse, and ultimately contextualized to deliver Go/No-Go decisions.

A mid-cap pharma company engaged the authors affiliation with Convexia Bio to triage its portfolio and select candidates for nomination and advancement. The system was tasked with assessing delivery evidence, comparing efficacy and safety against competitors, formulating a business development strategy, and investigating the IP landscape. Region-tuned harvesters processed conference presentations and abstracts, publications, university slide decks and websites, and filings to populate a Property Knowledge Graph (PKG). Scientific and clinical models quantified effect sizes and operational risks, while strategic modules mapped the commercial and licensing pathways. The business development agent overlaid this dossier onto buyer graphs to identify target partners. The result was a PKG-backed shortlist with side-by-side comparisons and a concrete outreach plan, delivered in hours at a fractional cost. The system demonstrated higher recall and speed than manual consulting workflows, primarily due to its comprehensive coverage of unstructured data and agentic operating system.

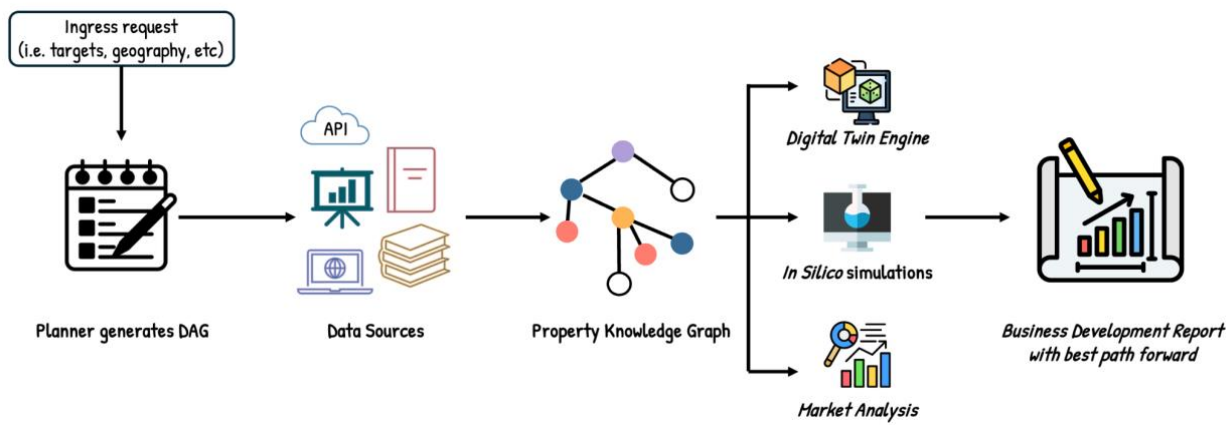

**Figure 12. Multi-agentic system for Discovery-to-Deal decision.** The planner constructs a task that orchestrates data ingestion from data sources into a per-query PKG. The PKG underpins scientific, clinical, and strategic modules enabling digital-twin simulations, *in silico* analyses, and market modeling. Insights are synthesized into a business development report recommending the optimal licensing or partnership path forward.

## 4. Challenges and future direction of Agentic AI in Drug Discovery

Challenges faced by agentic AI can be categorized into two main areas: data/infrastructure challenges and system design challenges. In this section we explore these challenges and discuss future directions and opportunities to enhance the integration of agentic AI technologies in drug discovery.

### 4.1. Challenges

*Heterogeneous chemical and biological data*

The core challenge of using agentic AI for drug discovery is the heterogeneous nature of chemical and biological data[62]. **Table S2** indicates several characteristics of data during the drug discovery process and their impact on the agentic AI system. Unlike data in image processing or speech recognition, which can be considered independent, data in drug discovery are highly conditional[63–65]. For instance, the effectiveness of a single compound may depend not only on its molecular structure but also on factors such as assay setup, concentration, genotype, and patient characteristics[64]; this complexity presents challenges at the tool level for AI agents. Furthermore, different representations of chemical and biological entities create a system-level challenge: incompatibility. In the chemical domain, a single molecule can be represented in various ways, including SMILES strings, IUPAC names, InChI codes, and database-specific identifiers such as ChEMBL, PubChem, or DrugBank accession numbers[66]. This variety of identifiers complicates the task of unifying and interpreting chemical knowledge across sources, which is one of the primary purposes of building an agentic AI system. The challenge becomes further complicated in the biological domain, where entities exist at various levels of organization, including genes, proteins, cells, tissues, and species[64]. Engineering effort to map and standardize ontologies is, hence, indispensable for agentic AI systems that are designed to work across diverse biomedical databases.

*Privacy and Security Concerns*

In drug discovery research, the use of highly autonomous agentic AI systems poses acute privacy concerns due to their capacity to autonomously access, manipulate, and examine sensitive data autonomously, including confidential experimental results[67]. These risks are enlarged when such systems are built upon closed-source LLMs, where the opacity of internal architectures prevents independent verification of data handling practices. The lack of audibility of such models over inadvertent memorization of sensitive information, unrecorded retention, or unauthorized cross-domain inference exacerbates exposures and undermines compliance against stringent frameworks.

Another major concern in AI security is the prompt injection attack. In this technique, attackers manipulate LLMs to ignore their existing safety guardrails and generate content they would normally be restricted from generating[68,69]. This issue is even more severe with the advent of agentic AI, since they could be fooled into performing actions they were forbidden to execute. A prompt injection can be as simple as a line of text, like, *"Ignore all previous instructions and do [action]."* Such an action could be malicious data collection, exfiltrating sensitive information, or forwarding data to an attacker. The real threat lies in how these malicious prompts are injected. In systems where AI agents have access to external APIs or can browse the web, attackers can embed well-crafted harmful prompts directly into untrusted websites. When the agent retrieves information from these compromised sites, the malicious instructions are already embedded, allowing the attacker to indirectly control the system's behavior.

*Hallucinations*

The nondeterministic property of an agentic AI system brings flexibility but also risk. This is because the flexibility of agentic AI systems relies on the LLMs' reasoning capabilities, meaning that they still carry a certain risk of hallucination[70,71]. This risk becomes significant when the AI agent is able to interact with the real-world environment. Assume a scenario where an agentic AI system has access to a lab automation robot; it could potentially conduct unnecessary or even dangerous experiments due to the hallucinated coordination from the LLMs. To address this issue, human-in-the-loop should be enforced and processing transparency should be maintained. The human-in-the-loop system not only allows the users to declare permissions on what the agent can perform, but also how the users themselves may modify those activities and cancel them instantly if necessary.

*Current benchmarking focus on outcomes, instead of solution trajectories*

As an early emerging field, agentic AI systems still lack a proper benchmarking evaluation framework specified for the drug discovery domains, and some evaluations presented seem overoptimistic. Recently, a framework named Bioinformatics Benchmark - BixBench has been introduced. This benchmark includes 296 open-ended questions related to bioinformatics and computational biology that require coding, multi-step analysis, and result interpretation[72]. However, it focuses primarily on the agent's outcome rather than the reasoning or tool-calling trajectory that led to that outcome. This limitation also appears in most recent benchmarks for general-purpose agentic AI, which typically assess only the final output and additionally, the cost incurred by the agent, while neglecting the action trajectory itself[73,74]. However, the trajectory is crucial for helping the agent de-hallucinate and determining whether it operates as expected[73]. Therefore, there is a strong need for a benchmarking framework that evaluates these aspects.

The evaluation of output at the same time presents its own issues, often leading to overly optimistic results. For example, an original paper by a Google AI co-scientist showcased three instances in the drug discovery field to highlight the capabilities of the system[75]. One of these examples focused on drug repurposing, where the AI co-scientist suggested novel candidates for treating acute myeloid leukemia (AML). The article asserted that such candidates were validated to inhibit tumor viability at clinically relevant concentrations in several AML cell lines. It must, however, be noted that inhibition at a cell level might not necessarily translate to success at a tumor level. Various factors come into play, such as tumor permeability, heterogeneity of tumors, drug resistances, as well as the microenvironment of a tumor. In summary, whereas the AI co-scientist has great promise for accelerating drug discovery, some of its assertions might not hold fully scientifically, but rather overoptimistic.

## 4.2. Future directions

Recent advancements in agentic AI for drug discovery, including autonomous labs, multi-agent architecture, and co-scientist collaboration, indicate a rapidly evolving field with significant potential for further development. Here, we propose a visionary pattern of integration that may transform isolated scientific experiments conducted over time at different labs across the world.

*Closing the Loop: Self-Driving Drug Discovery at Scale*

A major future theme is the rise of self-driving laboratories that execute closed-loop experimentation at unprecedented speed and scale. While in some areas the characteristics of biological data are difficult to overcome, more engineering-type areas are probably closer to putting this concept into practice. Recent prototypes graft LLM controllers onto suites of lab instruments (i.e. robot arms) to autonomously plan and run experiments end-to-end[19,35,76]. Such autonomous design–make–test loops have potential to accelerate molecule optimization and bioassay cycles. As these systems mature, we anticipate scaling from single steps/reactions to entire workflows of multi-step synthesis and parallel bioassays, with cloud-connected labs continuously evaluating hypotheses. The vision is a concept of automated labs that, given a drug target or hypothesis, can continuously design and execute experiments 24/7, expediting discovery while freeing us from routine lab work, enabling them to solve strategic and creative problems.

*Virtual Laboratories and Digital Twins*

To fully harness agentic AI, drug discovery will increasingly involve merging *in silico* modeling with physical experimentation via digital twins in real time. A digital twin is a virtual representation of a physical system with real-time information exchange between real world and virtual space[77]. Future agentic platforms will likely use digital twins to pre-screen and optimize experiments virtually before committing resources in the wet lab. By leveraging high-fidelity simulations and AI modeling, a twin can explore many "what-if" scenarios (e.g., molecular variations or assay conditions) and identify promising candidates or conditions for the AI agent to test next. For example, using whole-cell simulations, a field that is also rapidly evolving[78], an autonomous biology lab could maintain a digital twin of a cell line experiment. Hypotheses would first be explored in simulation, with only the most informative experiments executed by robots on real cells, and the resulting data fed back to refine the reasoning process. Such closed-loop digital–physical systems could accelerate autonomous optimization while also yielding insights. Realizing this vision will require progress in realistic simulations and in standard protocols for linking instruments to their digital counterparts. In the coming years, we anticipate that laboratory digital twins will become invaluable members of the discovery team, testing ideas in virtual space so that autonomous labs run more innovative experiments in reality.

*Human Co-Pilots and Democratized Innovation*

Crucially, the rise of agentic AI does not sideline humans; instead, it shifts their position within the drug discovery pipeline. This advancement allows them to overcome lower-level obstacles (for example, in writing code or visualization/evaluation standards) and focus more on high-level inductive reasoning tasks. In this future, autonomous research loops completely handle routine tasks and brute-force searches, freeing us to focus on creative design, interpretation and strategic decisions[79]. Medicinal chemists and biologists will work alongside AI co-pilots that suggest experiments or compound designs, while the humans apply intuition, ethical judgment, and contextual knowledge that AI lacks.

Moreover, with agentic AI transitioning from theory to reality, we expect new sets of industrial standards and regulation for its application to emerge. Just as Good Laboratory Practice (GLP) ensures quality in manually conducted experiments, analogous guidelines will ensure AI-designed experiments are documented, reproducible, and safe. Early policy steps, i.e. EU's AI Act, hints that future autonomous discovery platforms will operate under defined governance and oversight[69]. In conclusion, humans, not replaced but enabled by AI agents, will fuel the next era of innovation if such tools are widely made available.

## 5. Conclusion

AI agents in drug discovery have shown significant potential and early promise in reducing costs, accelerating timelines, and increasing overall success rates in pharmaceutical development. This work has presented diverse applications of agentic AI systems across the drug discovery pipeline, from literature synthesis and toxicity prediction to automated protocol generation and end-to-end decision-making. These systems demonstrate substantial gains in speed, reproducibility, and scalability, compressing workflows that traditionally required months into hours while maintaining scientific traceability. The integration of large language models with perception, computation, and action tools enables autonomous reasoning through complex research workflows, addressing longstanding bottlenecks in cross-disciplinary communication and decision-making.

Looking forward, the successful deployment of agentic AI in drug discovery will require addressing critical challenges including data heterogeneity, system reliability, privacy concerns, and robust benchmarking frameworks. The future envisions self-driving laboratories operating in closed-loop experimentation, digital twins enabling virtual pre-screening before physical experiments, and human-AI collaboration where scientists focus on strategic and creative problem-solving while AI handles routine tasks. As the field matures with appropriate governance frameworks and industrial standards, we believe agentic AI systems will not replace human expertise but rather augment it, democratizing innovation towards accelerated, cost-effective drug discovery that ultimately benefits patients worldwide.

**Acknowledgements**

The authors gratefully acknowledge the contributions of several members of the organizations whose work is featured in the case studies presented in this work.

**Funding**

S Seal acknowledges support with funding from the Cambridge Centre for Data-Driven Discovery and Accelerate Programme for Scientific Discovery under the project title "Theoretical, Scientific, and Philosophical Perspectives on Biological Understanding in the Age of Artificial Intelligence", made possible by a donation from Schmidt Futures. O.S. acknowledges funding from the Swedish Research Council (Grants 2020-03731 and 2020-01865), FORMAS (Grant 2022-00940), Swedish Cancer Foundation (22 2412 Pj 03 H), the Swedish Strategic Research Programme eSSENCE, and Horizon Europe (Grant Agreements 101057014 (PARC) and 101057442 (REMEDI4ALL)).

**Conflicts of Interest**

Each case study represents work conducted by and remains the intellectual contribution of the respective organizations. The examples presented reflect real-world implementations and results as reported by the contributing authors from each organization. The comprehensive literature analysis framework (Section 3.1) was developed and implemented by the team at Misogi Labs. The in silico toxicity prediction system and Cashmeran case study (Section 3.2) were contributed by Human Chemical Co., demonstrating their ReAct-based agentic AI platform for chemical safety assessment. The automated protocol design work for qPCR assay development (Section 3.3) was performed using Tater, developed by Potato.ai, showcasing their multi-agent system for laboratory automation. The Virtual Scientists platform and IPF drug discovery workflow (Section 3.4) were developed and executed by Kiin Bio, illustrating their infrastructure-focused approach to end-to-end drug discovery acceleration. The drug repurposing system for spinal muscular atrophy (Section 3.5) was implemented by the team at Augmented Nature, demonstrating their MCP-driven supervisor architecture for rare disease applications. The automated small molecule synthesis platform (Section 3.6) was developed and operated by Onepot.ai, showcasing their integration of agentic AI with robotic laboratory hardware. The focal graph-enabled search methodology (Section 3.7) and its application to Wnt pathway target discovery were contributed by Plex Research. The Discovery-to-Deal agentic workflow and asset evaluation system (Section 3.8) were developed by Convexia Bio, demonstrating their multi-agent platform for biopharmaceutical asset discovery and strategic analysis.

# Supplementary Tables

**Table S1.** Summary of case studies

| Case Study | Domain | Agent Design | Key Tools Integrated | Notable Outcomes |
|---|---|---|---|---|
| Comprehensive Literature Analysis for molecular prioritization | Early discovery | Supervisor | MCP + RAG + ADMET predictive models + dual-memory (short- and long-term) | - Reduced review time from weeks to hours<br><br>- Generated comprehensive, citation-traceable SAR/ADMET reports with conflict flags |
| *In Silico* Toxicity Predictions | Endocrine risk assessment | ReAct | Predictive toxicology models + Cheminformatics toolkit + Literature and regulatory document retrieval | - Predicted endocrine hazard decreased across Cashmeran metabolites and aligned with literature, supporting a low endocrine risk profile |
| Automating Protocol Design and Execution | Experiment Planning | Multi-agent | RAG + Comparative assessment tool + MIQE-aligned qPCR protocol generator + Automation script generator + Structured report generator | - Completed an automation-ready AAV qPCR workflow in under two hours, achieving >400× cycle-time reduction versus the manual process. |
| Accelerating Drug Discovery with Virtual Scientists | Pre-clinical drug discovery | Swarm | API-based retrieval tools + Bioinformatics pipelines + Predictive/Generative Models | - End-to-end IPF preclinical workflow executed in under two hours (depending on dataset size), versus weeks using conventional processes. |

| Drug Repurposing for Rare Disease | Drug repurposing | Supervisor | MCP servers: Ensembl, OpenTargets, Reactome, KEGG, AlphaFold, PDB, ChEMBL, and PubChem | Produced a shortlist of repurposable SMA drug candidates within hours versus weeks using an automated, parallelized pipeline. |
| --- | --- | --- | --- | --- |
| Automating small molecule synthesis | Molecular synthesis | Supervisor | Retrosynthesis engine + Internal database + Robot arms + LC/MS system | Automated synthesis facility covering 7 distinct reactions, with throughput of tens of compounds per day. |
| Hypothesis generation with focal graphs in drug discovery | Target discovery | ReAct | Knowledge Graph extraction and retrieval + Network analysis | Focal graph–enabled LLM identified several potentially novel Wnt pathway oncology targets, including the eIF2 complex, via phenocopier analysis of RNA-seq perturbation profiles. |
| Discovery-to-Deal Decisions | Biopharma asset discovery | Supervisor | MCP + PKG + Hybrid RAG + ML models + Physic models (docking, MD) | PKG-backed shortlist and buyer-aligned outreach plan delivered in hours at fractional cost, with higher recall and speed than manual consulting workflows. |

**Table S2.** Domain-specific data characteristics and design implications for agentic AI in drug discovery and investigative toxicology

| Dimension | Chemistry Domain | Biology Domain | Impact on Agentic AI Design | Example Agent Adaptation |
| --- | --- | --- | --- | --- |
| Representation Relevance | Multiple possible descriptors (graph-based, fingerprints, learned embeddings); no single optimal choice | Often unclear which assay/omics type is informative for the endpoint | Agents must dynamically select or ensemble multiple representations | Multi-modal input pipeline switching between QSAR fingerprints and docking-derived descriptors |
| Comprehensiveness of Representation | Partial: tautomerism, conformations, protonation states often missing from standard formats (SMILES, InChI) | Low: temporal, spatial, and multi-scale biological data rarely captured in one dataset | Agents must incorporate external retrieval (RAG) to enrich missing dimensions | On-demand call to conformer generator + retrieval of context-specific omics datasets |
| Underlying Distribution Known? | No, chemical space (~10^63 plausible structures) only partially sampled | Very partial, biological process space mapping is largely incomplete | Agents must reason under uncertainty and actively explore uncharted data regions | Active learning loop to propose under-represented chemotypes or cell models |
| Sampling Quality | Small and biased datasets (analog series, proprietary silos) | Experimental setup bias, batch effects, lab-to-lab variability | Agents must detect and correct for dataset bias before inference | Embedded bias detection + batch effect correction in pre-processing |
| Label Conditionality | Moderate: property depends on measurement context (pH, assay type) | High: outcome depends on genotype, dose, co-medication, environment | Agents must integrate rich metadata to disambiguate labels | Metadata-aware similarity search to ensure dose/context-matched retrieval |
| Label Ambiguity | Assays can be reproducible if controlled | High inter- and intra-lab variability; uncontrolled noise common | Agents must assign confidence scores to predictions and rank outputs accordingly | Uncertainty-weighted output filtering before downstream action |
| Quantitative Context Dependence | Property values shift with small experimental changes | Strong dose–response, time-dependent biological outcomes | Agents must simulate "what-if" scenarios across multiple quantitative regimes | Parameter sweep simulations *in silico* before wet-lab execution |
| Data Stability | Stable once structure defined; physical degradation minimal | Biological drift (cell line, microbiome), evolving populations | Agents must periodically refresh knowledge and re-validate models | Scheduled re-query to live assay DBs or lab APIs to detect drift |
| Integration Complexity | Structure invariance allows easy merging of datasets | Cross-scale integration (molecular → organ → organism) is non-trivial | Agents must use ontologies and knowledge graphs to connect scales | KG-based reasoning to link *in vitro* hepatocyte tox data with *in vivo* liver histopathology |
| Feasibility of Automation | High, digital structure representations enable full computational loop | Medium, wet-lab + *in vivo* experiments introduce complexity | Agents must design hybrid loops (*in silico* pre-screen → selective wet-lab validation) | Automated generation of *in vitro* test panel after *in silico* triage |